\documentclass[conference]{IEEEtran}
\IEEEoverridecommandlockouts
\usepackage{times}
\usepackage{latexsym}
\usepackage{url}

\usepackage{hyperref}
\RequirePackage{natbib}

\usepackage{makecell}
\usepackage{multirow}

\usepackage{natbib}
\setcitestyle{square,numbers}

\usepackage{graphicx}

\usepackage{enumitem}

\usepackage{booktabs}
\usepackage{siunitx}
\usepackage{etoolbox}
\robustify\bfseries

\usepackage{xspace}
\makeatletter
\DeclareRobustCommand\onedot{\futurelet\@let@token\@onedot}
\def\@onedot{\ifx\@let@token.\else.\null\fi\xspace}

\def\eg{e.g\onedot} 
\def\ie{i.e\onedot}

\makeatother

\begin{document}

\title{A Survey of Methods to Leverage Monolingual Data in Low-resource Neural Machine Translation}

\author{
\IEEEauthorblockN{1\textsuperscript{st} Ilshat Gibadullin}
\IEEEauthorblockA{\textit{Innopolis University}\\
Innopolis, Russia \\
i.gibadullin@innopolis.ru}
\and
\IEEEauthorblockN{2\textsuperscript{nd} Aidar Valeev}
\IEEEauthorblockA{\textit{Innopolis University}\\
Innopolis, Russia \\
ai.valeev@innopolis.ru}
\and
\IEEEauthorblockN{3\textsuperscript{rd} Albina Khusainova}
\IEEEauthorblockA{\textit{Innopolis University}\\
Innopolis, Russia \\
a.khusainova@innopolis.ru}
\and
\IEEEauthorblockN{4\textsuperscript{th} Adil Khan}
\IEEEauthorblockA{\textit{Innopolis University}\\
Innopolis, Russia \\
a.khan@innopolis.ru}
}

\maketitle

\begin{abstract}
Neural machine translation has become the state-of-the-art for language pairs with large parallel corpora. However, the quality of machine translation for low-resource languages leaves much to be desired. There are several approaches to mitigate this problem, such as transfer learning, semi-supervised and unsupervised learning techniques. In this paper, we review the existing methods, where the main idea is to exploit the power of monolingual data, which, compared to parallel, is usually easier to obtain and significantly greater in amount.
\end{abstract}

\section{Introduction}

A lack of parallel data is a major problem of machine translation for many language pairs. This is usually the case when one or both languages in a pair have a small number of speakers or low media presence. However, if there's some parallel data, then, typically, there exists orders of magnitude more monolingual data, which, in addition to parallel one, can result in significant improvements in translation quality. 

The success of neural networks in machine translation task motivates the exploration of methods for efficient application of monolingual data over them. Over the last few years a lot of work has been done in this direction, and many new approaches have been suggested \cite{gulcehre,sennrich,cheng,zhang, currey,domhan,stahlberg, ramachadran, skorohodov}. If someone is looking for a way to benefit from monolingual data, one needs to go through all these works, to be able to understand, compare, and choose. However, if these methods were grouped by some similarity criteria and better organized, then navigating through them would be substantially simplified, both for practical use and for research needs. We have seen some surveys for other subfields of machine translation, e.g., a survey of domain adaptation techniques \cite{chu2018survey} or a study of post-editing methods \cite{koponen2016machine}, but to the best of our knowledge, there are no surveys of approaches which utilize monolingual data. In this paper, we attempt to solve this problem by reviewing, categorizing, and comparing the existing methods of exploiting monolingual data in neural machine translation (NMT) models. This work is meant to be a good starting point for a quick immersion into the subject.



\subsection{NMT Models}

All methods, which will be covered, are based on a couple of fundamental NMT models. Both models consist of encoder and decoder parts, where the role of the encoder is to represent an input sequence as a context-dependent vector, while the purpose of the decoder is to generate the sequence in the target language based on the encoder output. The first model was introduced by \citet{bahdanau} and is based on recurrent neural network (RNN) layers for encoder-decoder parts of the model and attention layer between them. Further in the paper, we will refer to this model as RNN-based. The second model was presented by \citet{vaswani} and is called Transformer. They introduced a new technique called multi-head attention, and encoder-decoder parts of the model are based on stacks of multi-head attention and feed-forward layers. Such architecture allows to get rid of RNN layers, so the training becomes significantly faster.

\subsection{Organization of Methods}

We divided all existing methods into two broad categories: \textit{Architecture Independent} and \textit{Architecture Dependent} methods. Such division is made from the practical viewpoint: there exist several strong NMT models and development of new models or some modifications to existing ones continues; so, Architecture Independent methods of exploiting monolingual data may be used with any model to get translation quality improvements, viewing a model as a black box. On the contrary, Architecture Dependent methods require specific changes in architecture and may or may not be adapted to various other NMT models.
\par
We can relate to \textbf{Architecture Independent} methods all approaches, the idea of which is to generate pseudo-parallel corpus using monolingual corpus, then mix pseudo-parallel corpus with true parallel corpus and make no distinction between them during training. Also, we can include to this category those methods which use separate pre-trained language model and merge it with pre-trained translation model during inference.
\par
\textbf{Architecture Dependent} methods focus on specific architectural features of NMT models and/or require additional changes in architecture. One type of methods only freezes some parameters of NMT model during training on pseudo-parallel corpus. Other methods apply unsupervised pre-training, using monolingual corpus, of some parameters of NMT model, then initialize it with these pre-trained parameters to further train on parallel corpus. More complex methods integrate language modeling idea and/or multi-task learning to NMT model. 

There is a recent trend towards using monolingual data in fully \textbf{unsupervised} setting. As this approach is totally different from other methods we describe here (which use monolingual data only in addition to the parallel one), we do not include it in our taxonomy, instead describing in separately.

\begin{figure*}
\centering
\includegraphics[width=0.8\linewidth]{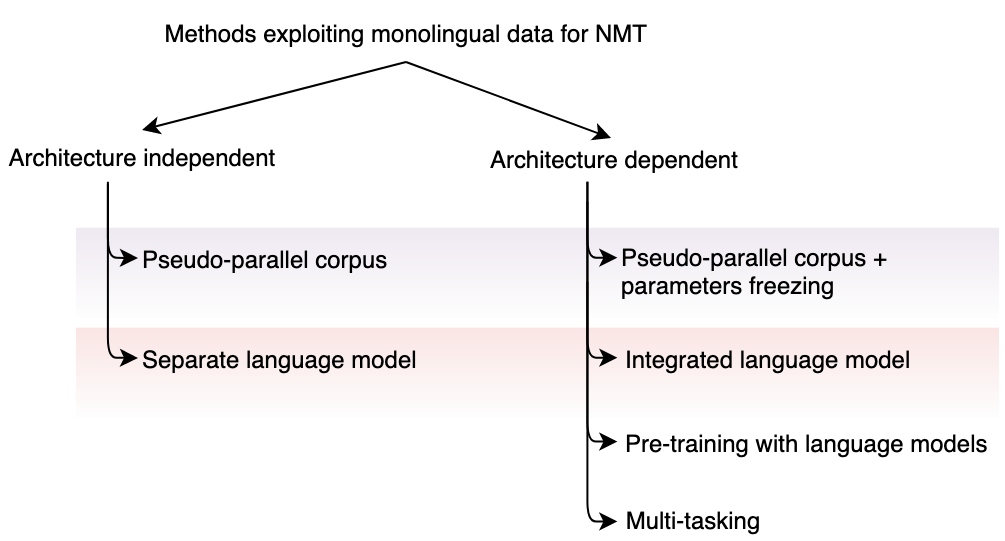}
\vspace*{+5mm}
\caption{The taxonomy of methods exploiting monolingual data for NMT. The shadowed areas highlight the categories sharing the same core idea.}
\label{fig:methods}
\end{figure*}

The taxonomy of methods is presented in Figure \ref{fig:methods}. Below, we will provide further categorization with examples and results overview for each method.

\section{Architecture Independent Methods}
Architecture independent methods of exploiting monolingual data are transparent to NMT model architecture, so any existing model may be used as a base. We will consider methods from this category divided into two subcategories according to the way monolingual data is used:
\begin{enumerate}
\item Methods which use additional pseudo-parallel corpus;
\item Methods which merge NMT model with a separate language model.
\end{enumerate}

\subsection{Additional Pseudo-parallel Corpus}
\label{Additional pseudo-parallel corpus}

The main idea of methods from this subcategory is to generate pseudo-parallel (or synthetic) corpus using monolingual data so that we have additional input and output for our NMT model. Then we mix pseudo and real parallel corpora and make no distinction between them during training. We can generate either input or output part from the existing target or source monolingual sentences of some language pair. The benefit of pseudo-parallel corpus is that the NMT model will better learn the structure of target or source language, depending on the side of monolingual data. The drawback is that the low quality of generated sentences or domain mismatch may degenerate the learned structure of the corresponding side, so in some cases we have to limit the size of pseudo-parallel corpus. Further, we will consider methods to generate a pseudo-parallel corpus.

\subsubsection{Back-Translation}
\label{Back-Translation}

This idea of synthetic parallel corpus generation was proposed by \citet{sennrich} and has the following idea: an additional reversed machine translation (MT) model is trained on available parallel corpus in target-to-source direction, \ie target and source sides with respect to the main translation model. This opposite direction is important because thereby the target side stays intact. Then, using this pre-trained reversed model target monolingual sentences are translated to the source language. These sentence pairs form a new pseudo-parallel corpus. Then, true and synthetic parallel corpora are mixed, and the main model is trained on this combination. 
\par
This method was evaluated on low-resource English\(\rightarrow\)Turkish language pair and RNN-based NMT model was used as a baseline. 300K of parallel sentence pairs and 3.2M of synthetic parallel sentence pairs, back-translated from monolingual sentences, were used for training. The addition of synthetic training data led to an improvement of 2.7 BLEU on average. The authors also analyzed how final NMT model's quality depends on the quality of back-translated sentences and found out that the quality improvement of back-translated sentences by 6 BLEU gives 0.6 BLEU improvement of the final NMT model. Experiments conducted by \citet{stahlberg} show that performance of Back-Translation doesn't degrade as long as the ratio of synthetic to real parallel corpora doesn't exceed 8:1 ratio.

\subsubsection{Round Trip Training}
\label{Auto-encoders}

This method doesn't generate pseudo-parallel corpus explicitly as the previous method. Instead, it leverages the idea of auto-encoders to generate pseudo-parallel sentence and immediately reconstruct it back.
\par
Simply speaking, the main purpose of auto-encoders in deep learning is to learn the common input features. It has two parts called encoder and decoder. The role of the encoder is to extract the common input features, and the role of the decoder is to reconstruct the input from encoder's output.
\par
The method proposed by \citet{cheng} uses auto-encoders to exploit monolingual corpora. The idea is as follows. There are two NMT models, first with source-to-target and second with target-to-source translation directions respectively. Source-to-target NMT model can be considered as an encoder and target-to-source NMT model as a decoder part of the auto-encoder network, which role is to reconstruct source sentences. The same auto-encoder may be constructed in the opposite direction, where target-to-source model is considered as an encoder part and source-to-target model as a decoder part of the auto-encoder network, which role is to reconstruct target sentences. The whole training objective of the method is to maximize the likelihoods of source-to-target and target-to-source models on parallel corpus, and reconstruction likelihoods of auto-encoders on monolingual corpora. On each iteration, models are trained by mini-batch of parallel corpus and two mini-batches of target and source monolingual corpora. 
\par
This method can be seen as an iterative extension to Back-Translation and may exploit not only target-side but also source-side monolingual data.
\par
The authors evaluated the method on Chinese\(\leftrightarrow\)English language pair and used RNN-based NMT for all experiments. 2.56M parallel sentence pairs, 18.75M Chinese and 22.32M English monolingual sentences were used for training. The authors found out that using both source and target monolingual data doesn't give significant improvements over baseline. Using parallel and English monolingual corpora, the authors achieved +4.7 BLEU over baseline in Chinese\(\rightarrow\)English direction. The same result with parallel and Chinese monolingual corpora in English\(\rightarrow\)Chinese direction was obtained. The method also outperforms Back-Translation by +1.8 and +1.0 BLEU in Chinese\(\rightarrow\)English and English\(\rightarrow\)Chinese directions, respectively. We think that this may happen because with each new iteration the translation quality improves; thus, the quality of synthetic sentences also becomes higher, while in Back-Translation the quality of synthetic parallel sentences is constant. Using source-side monolingual data also gives improvements over baseline and Back-Translation, but smaller than those, obtained with target-side monolingual data.

\subsubsection{Copied Monolingual Data}

This method explicitly generates pseudo-parallel corpora, but in contrast to Back-Translation, no additional translation model is used.
\par
The method proposed by \citet{currey} suggests converting target-side monolingual data to synthetic parallel corpus by copying target-side sentences to the source side. To represent source and target words in the same vocabulary, they use byte-pair encoding \cite{sennrich2}. Parallel corpus is mixed with synthetic parallel corpus, and no distinction is made during training.
\par
The authors consider this method as a multi-task system \cite{luong, johnson}, where one NMT model combines several translation directions, e.g. French\(\rightarrow\)English, German\(\rightarrow\)English, and English\(\rightarrow\)German. This method combines \eg English\(\rightarrow\)English and Turkish\(\rightarrow\)English into one system for the purpose of improving Turkish-to-English quality.
\par
The evaluation was performed on English\(\leftrightarrow\)Turkish language pair, and RNN-based model was used for experiments. 207K parallel sentence pairs, 414K English and 414K Turkish monolingual sentences were used for training. The method gave +1.2 BLEU over baseline. Increasing the ratio of copied to parallel text was found to improve BLEU score: 3:1 ratio gives +0.8 BLEU over 1:1 ratio for English-Turkish pair. However, additional experiments with an increased ratio of copied to parallel text are required, because the model will most likely start to degrade with higher ratios.

\subsection{Separate Language Model}
\label{Separate language model}

In this subcategory, we will describe methods which use an additional separate target-side language model (LM). LMs are pre-trained on target-side monolingual corpus to predict next word probability distribution based on given previous words \(P_{LM}(y_t|y_1^{t-1})\).  The main idea of these methods is to merge LM and translation model (TM) in the prediction step. This means that the probability distribution of next word given previously predicted words and input sentence \(P(y_t|y_1^{t-1}, x)\) is calculated using output logits (or probability distribution) of TM and output logits of LM. The expected effect of additional target-side LM is that it should help TM to generate grammatically correct sentences.
\par 
As LM is independent of TM, there're no restrictions on its architecture: it can be e.g., n-gram based feed-forward LM \cite{bengio} or RNN-based LM \cite{mikolov}.

\subsubsection{Shallow Fusion}

Shallow Fusion \cite{gulcehre} is a technique to merge separately pre-trained LM and TM. Each time step (step of prediction of next word) TM and LM propose scores of next possible words based on previously predicted words. Then, the scores of TM are summed with the scores of LM multiplied by hyper-parameter \(\beta\), which requires additional fine-tuning on validation data, to control LM influence. The word with the highest score is selected to be the next word in a sequence. To be more accurate, beam search is applied, where top N most probable sequences are carried until the end of prediction and the most probable one (which has the highest score) is chosen as a result.
\par
The evaluation is provided for Turkish\(\rightarrow\)English translation direction, and all experiments used RNN-based TM and LM. 160K parallel sentence pairs and 4M English monolingual sentences were used for training. Unfortunately, no improvements were obtained using this method, it gives almost the same results as a baseline without LM.

\subsubsection{PostNorm}

This method uses the same idea of merging technique from the previous method with some modifications. The main difference here is that LM is trained first, then merged with TM, then, fixing LM's parameters, TM is trained. This technique is inspired by Cold Fusion \cite{spiram}.
\par
This LM and TM merging technique is called PostNorm and was proposed by \citet{stahlberg}. First, the output of projection layer \(S_{TM}(y_t|y_1^{t-1}, x)\) of TM is normalized by softmax and the probability distribution \(P_{TM}(y_t|y_1^{t-1}, x)\) is obtained. Then, it is component-wise multiplied by probability distribution of LM \(P_{LM}(y_t|y_1^{t-1})\). To make it a valid probability distribution, additional normalization using softmax function is applied. In contrast to Shallow Fusion, this method requires normalization after the summation of log probabilities (multiplication of probabilities) and training of TM after merging with LM.
\par
The method was evaluated on Turkish\(\leftrightarrow\)English language pair, RNN-based TM and LM were used in experiments. 207K parallel sentence pairs, and 47M English and 3M Turkish monolingual sentences were used for training. The method gives +0.71 BLEU for English\(\rightarrow\)Turkish and +0.43 BLEU for Turkish\(\rightarrow\)English directions over baselines without LM. For higher-resourced Xhosa\(\rightarrow\)English pair with 739K parallel sentences, it achieves +2.36 BLEU.

\section{Architecture Dependent Methods}

In contrast to architecture independent methods, which are applicable to any existing NMT model, architecture dependent methods are designed for specific NMT model and may require significant modifications of its architecture. Similarly to the previous category, we divide the methods here in subcategories by their approach to using monolingual data:
\begin{enumerate}
\item Methods which freeze some parameters during training on pseudo-parallel corpus;
\item Methods which integrate language modeling in NMT model architecture;
\item Methods which pre-train parts of NMT model with language models;
\item Methods which use monolingual data for multi-tasking.
\end{enumerate}

\subsection{Training with Parameters Freezing}

The idea of methods in this subcategory is the same as in \ref{Additional pseudo-parallel corpus}, except that we distinguish between pseudo-parallel and real-parallel corpora, and freeze some trainable parameters of NMT model during training on pseudo-parallel corpus. Parameters freezing operation is introduced to weaken the negative influence of generated sentences to the corresponding side of NMT model. However, we can't completely overcome the negative effect of sentences with low quality; thus, we still need to limit the size of pseudo-parallel corpus. Both methods from this subcategory use RNN-based NMT model.

\subsubsection{Forward-Translation}
\label{Forward-translation}

This method by \citet{zhang} is similar to Back-Translation (\ref{Back-Translation}), except that here source-side monolingual corpus is exploited and parameters freezing is used during training on synthetic parallel corpus.
\par
As in the Back-Translation method, any NMT or SMT model is trained as an additional TM to generate synthetic corpus by translating monolingual one, but from source to target (forward translation). Further, parallel and synthetic data are combined together to train the main translation model. To protect the decoder part of the model from the negative influence of synthetic corpus, the authors freeze decoder parameters during training on synthetic corpus.
\par
The method was evaluated on Chinese\(\rightarrow\)English translation direction, and RNN-based NMT model was used for forward translation. Using 620K parallel sentences and 3.3M Chinese monolingual sentences, the method outperforms baseline NMT model by 3.2 BLEU.

\subsubsection{Dummy Input}

This method was proposed by \citet{sennrich} together with Back-Translation. The idea of the method is the following: target-side monolingual sentences are paired with single-word dummy \(<\)\(null\)\(>\) input to produce pseudo-parallel corpus, which is used for NMT model training. To protect encoder and attention parts of RNN-based NMT model from low-quality input, the parameters of encoder and attention layers are fixed during training on this pseudo-parallel corpus.
\par
The authors justify the efficiency of this method saying that it allows to better learn the target language structure, but significantly harms the conditioning of decoder part of the model on encoder and attention layers if the ratio of monolingual data is too high. The advantage of this method compared to Back-Translation is reduced time to fit the system, as there's no need in training of additional NMT model and in Back-Translation operation, both of which are quite time-consuming. 
\par
Turkish\(\rightarrow\)English translation direction was used for evaluation with 300K of parallel sentences and 177M of English monolingual sentences. On average, there is an improvement of 0.6 BLEU over baseline if monolingual corpus with dummy source-side is added to parallel data in 1:1 ratio. Higher proportions of dummy source-side degrade the BLEU score.

\subsection{Integration of Language Modeling}

In contrast to the methods described in \ref{Separate language model}, where we merged language model with translation model in prediction step, in this subcategory we will cover methods, which integrate language modeling technique to NMT model architecture, so merging of LM with TM occurs in earlier stages. The benefit of such integration is that we can more accurately exploit LM, considering architectural features of TM. Both methods from this subcategory are developed for RNN-based NMT model.

\subsubsection{Deep Fusion}

This method was proposed by \citet{gulcehre} with Shallow Fusion technique and also uses separately pre-trained RNN-based LM and TM, but the difference is that merging of LM with TM occurs earlier.
\par
In this method, called Deep Fusion, the hidden state of LM is concatenated with the decoder's hidden state. Thus, the new hidden output will be added to the decoder's input: LM's state in addition to TM's state, embedding of the previous word and the context vector. To balance the influence of LM on TM, the additional controller mechanism was proposed to control the magnitude of LM's hidden state. Then, the model's hidden output and controller mechanism parameters are fine-tuned on training data.
\par
For Turkish\(\rightarrow\)English translation direction with 160K parallel sentences and 4M English monolingual sentences the method gives on average +1.19 BLEU over baseline without LM.

\subsubsection{Language Model with Multi-Task Learning}
\label{Language Model with Multi-Task Learning}

This method differs from the previous one by that it doesn't use pre-trained LM and TM, and has a different way of integrating RNN-based language model \cite{mikolov}.
\par
The method proposed by \citet{domhan} uses an additional source-independent RNN layer, which has the language modeling role. Source-independent means that the parameters of this layer will only be affected by target-side sequences. To compute its current hidden state, this RNN layer takes as an input its previous state and embedding of the previous target word. Computed current hidden state of this layer goes as an input to the decoder layer of RNN-based model, instead of embedding of the word, to compute the decoder's hidden state. Here is the difference from Deep Fusion, where we just concatenate the hidden states of LM and TM's decoder. To jointly train LM and TM, the second objective is specified for the output of the language modeling layer---to predict the next target word, conditioned only on target history information. The system is trained to minimize the loss of the model by updating source-dependent parameters only on parallel corpus and updating source-independent parameters on both monolingual and parallel corpora.
\par
The evaluation is provided for English\(\rightarrow\)German and Chinese\(\rightarrow\)English translation directions. 191K English-German and 242K Chinese-English parallel sentences pairs, 51M German and 51M English monolingual sentences were used for training. The method got +1.4, +0.5 BLEU improvements for English\(\rightarrow\)German and Chinese\(\rightarrow\)English directions respectively over baseline without LM. The authors also provided a comparison with Back-Translation method \cite{sennrich}, which outperforms on average by +3 BLEU. The difference may be explained by the fact that the gradients from monolingual data don't change the source-dependent part of the model; in contrast, synthetic data always affects all model parameters. The authors think that training with synthetic source data may also act as a model regularizer.

\subsection{Pre-training with Language Models}
\label{Pre-training}

Here we will describe one more subcategory of methods, which exploit the power of language modeling. Pre-training with language models is a technique when parts of NMT model are trained as LMs, and parameters of these pre-trained parts are used to initialize NMT model. We will describe two methods, which use the same idea but apply it for two different NMT models: RNN-based model \cite{bahdanau} and Transformer \cite{vaswani}. Both models are based on encoder-decoder architecture, where encoder part may be considered as a source-side LM and pre-trained on a source-side monolingual corpus, and decoder part may be considered as a target-side LM and pre-trained on a target-side monolingual corpus. The benefit of pre-training with LMs is that we can efficiently exploit both source-side and target-side monolingual corpora.

\subsubsection{Pre-training of RNN-based model}

In this method \cite{ramachadran}, source-side LM is the encoder part of RNN-based model with an additional softmax output layer. Target-side LM is the decoder part of RNN-based model without encoder input. Both LMs are pre-trained on corresponding monolingual corpora. After pre-training, the embedding and first RNN layers of encoder and decoder, plus decoder's softmax layer, are initialized with pre-trained weights. Then, the model is fine-tuned using parallel corpus. To ensure that the model doesn't overfit on labeled data forgetting language modeling task, the authors continued training on monolingual data, setting equal weights to LM losses and translation model loss. Also, the authors improved the model by adding a residual connection between the first RNN layer of decoder and softmax layer, because the intermediate decoder layers are randomly initialized and thus may break LM of target-side by random gradients.
\par
Evaluated on English\(\rightarrow\)German translation direction using WMT 14 dataset with 4 million parallel and monolingual sentences, the method outperformed Back-Translation (\ref{Back-Translation}) by 0.3 BLEU.

\subsubsection{Pre-training of Transformer model}

Similarly to the previous method, \citet{skorohodov} train source-side and target-side LMs, then use all pre-trained LM parameters to initialize NMT model. Connections from the hidden state of encoder to logits are discarded and attention weights from encoder outputs to decoder are  randomly initialized. In Transformer, a decoder layer stack consists of the following layers: self-attention layer, encoder-decoder attention layer (where encoder output comes), and feed-forward layer. The encoder-decoder attention layer is initialized randomly, so to protect the target-side LM from breaking by random gradients, an additional residual connection between self-attention layer and feed-forward layer was introduced.
\par
This method was examined on extremely low size parallel corpus. The evaluation on Russian\(\rightarrow\)English translation direction with only 20K parallel sentences and 500K English sentences showed +1.4 BLEU improvement over Transformer without LMs initialization.

\subsection{Multi-tasking}
In this subcategory, we will describe only one method, which uses monolingual data in a different way---neither translating it nor using LMs.

\subsubsection{Input Sentences Reordering}
 Methods to use source-side monolingual corpus were described in \ref{Auto-encoders}, \ref{Forward-translation}, \ref{Pre-training}, but because of the simplicity of the models, monolingual data may not have been used to its full potential. The idea of this method is to use a more complex model to more efficiently leverage source-side monolingual data. The method is based on sentence reordering technique, which tries to reorder words in source-side language sentence so as to approximate the target-side language words order. In all experiments, RNN-based NMT model was used.
\par
The method proposed by \citet{zhang} has the following idea: there's a single shared encoder and two different decoders. The first decoder is used for translation, the second for reordering, where the target is just a reordered source sentence. Target sentences for training were obtained from source sentences by applying reordering rules proposed by \citet{wang}. The objective of the whole model is to maximize the sum of log probabilities of translation and reordering models. Training is performed by iterations of 5 epochs. One epoch for training reordering model, others for training translation model. The method applies multi-tasking because it performs translation and reordering at the same time.
\par
The evaluation is provided for Chinese\(\rightarrow\)English direction with 620K parallel sentences and 6.5M Chinese monolingual sentences. The method gives +4.3 BLEU over baseline, which is a considerable improvement.

\section{Fully Unsupervised Learning}

Unlike all the methods we described above, this group of approaches doesn't treat monolingual data as an additional resource for improving NMT models trained on parallel data. Instead, models here are trained using only monolingual corpora. This is made possible due to linguistic similarities between source and target languages.

The first step in unsupervised learning is bilingual dictionary induction---having it, one can already build a simple word-by-word translation model and then improve it.

There are several approaches to build such dictionary. Initially, the bilingual dictionary generation method was suggested by \citet{mikolov2013}. The authors found out that word embeddings of two different languages can be mapped to a common space where words from different languages with the same meaning reside close to each other. If the first language embeddings matrix is \(X\) and the second language embedding matrix is \(Y\), then one can find such linear transformation matrix \(W\) that \(WX\) is in the same space as \(Y\). In this common space, translations of words in one language can be found by searching nearest neighbors among the words from another language. The transformation matrix \(W\) can be found using some small seed dictionary or even without it, as it was suggested by \citet{artetxe2018acl}.

When common embeddings space is found, a simple word-by-word translation model can be built and iteratively improved using such techniques as language modeling based on denoising auto-encoders, iterative back-translation, etc.. The detailed description of such methods (\cite{artetxe2018unsup}, \cite{lample2018newest}) is out of the scope of this survey.

The results obtained by unsupervised methods are impressive. Here we provide the ones demonstrated by \citet{lample2018newest}. The method was evaluated on English\(\leftrightarrow\)French language pair using News Crawl WMT14 monolingual corpora. The reported scores are 25.1 and 24.2 BLEU in English\(\rightarrow\)French and French\(\rightarrow\)English directions respectively, using Transformer model. The authors performed a comparison with a model trained on parallel data, varying the number of parallel sentences, and found out that their method obtains higher BLEU score as long as the model trained on parallel data uses less than 300-400K parallel sentence pairs. For example, when 100K parallel sentence pairs are used, their method outperforms by +2.6 BLEU.

\begin{table}
\begin{center}
\begin{tabular}{|l|l|l|l|}
\hline \bf Model & \bf \(<\)100K & \bf 100-300K & \bf \(>\)300K \\ \hline
\multicolumn{4}{|l|}{\textit{Architecture independent}} \\ \hline
Back-Translation & - & 2.7 & - \\ \hline
Round Trip Training & - & - & 4.7 \\ \hline
Copied Monolingual Data & - & 1.2 & - \\ \hline
Shallow Fusion & - & 0 & - \\ \hline
PostNorm & - & 0.57 & 2.36 \\ \hline
\multicolumn{4}{|l|}{\textit{Architecture dependent}} \\ \hline
Forward-Translation & - & - & 3.2 \\ \hline
Dummy Input & - & 0.6 & - \\ \hline
Deep Fusion & - & 1.19 & - \\ \hline
LM with Multi-Task learning & - & 0.95 & - \\ \hline
Pre-training of RNN-based model & - & - & - \\ \hline
Pre-training of Transformer & 1.4 & - & - \\ \hline
Input Sentences Reordering & - & - & 4.3 \\ \hline
\textit{Fully Unsupervised Learning} & 17-2.6 & 2.6-0 & 0 \\ \hline
\end{tabular}
\end{center}
\caption{\label{bleu_comp} Positive BLEU improvements of covered models depending on size of parallel corpus used for training.}
\end{table}

\section{Comparison}
The common issue with the reviewed models is that the majority of them are evaluated on different datasets, so it's difficult to compare their effectiveness. However, some trends can be observed across the models, and we discuss them below. Results of all methods we collected in Table \ref{bleu_comp}.

The results obtained using Architecture Independent methods show that approaches which use Additional Pseudo-parallel Corpus have significant improvements over baseline NMT model. Especially high scores were obtained for Round Trip Training method, but the evaluation was performed on 2.56M of parallel sentences, so the additional experiments on less parallel data are required to prove its effectiveness for low-resource MT. Back-Translation shows a good result of +2.7 BLEU using 300K parallel sentences. Methods exploiting Separate Language Model, however, don't obtain such high scores. Nevertheless, methods from these two subcategories may be applied together to get even better results.
\par
Among Architecture Dependent methods, Input Sentence Reordering shows the exceptional result of +4.3 BLEU over baseline, evaluated on 620K of parallel data. Forward-Translation, a technique which uses Additional Pseudo-parallel Corpus and parameters freezing, also gives a high score of +3.2 BLEU over baseline on the same parallel data. In contrast to Architecture Independent methods, Architecture Dependent methods of leveraging language models show better results when evaluated on small parallel corpora. Deep Fusion, a method to integrate language model into NMT architecture, gives +1.19 BLEU for low-resource language pair with 160K parallel sentences. Another approach, where Transformer model is pre-trained with language models, outperforms baseline Transformer by +1.4 BLEU using only 20K parallel sentences. 
\par

\section{Open Research Questions}
Monolingual data can be characterized by its amount and domain. \citet{stahlberg} demonstrate that for Back-Translation increasing the amount of monolingual data improves the translation quality only up to some point, and then it starts to degrade. \citet{zhang} also show that the partial use of monolingual data is beneficial, but in their case, they use the most relevant part of it. \citet{gulcehre} further highlight the importance of language domain.

Some questions which remain unanswered are: 1) How to determine the optimal proportion of monolingual data when it comes from the same or different domain? 2) What are the effective domain adaptation techniques in case of using language models and in case of using raw monolingual data? 3) How to use monolingual data to its full potential, such that a model can benefit from out-of-domain data too?  

It is not always possible to acquire monolingual data in the same domain as parallel data, especially for low-resource language pairs, therefore answering the above questions can be highly beneficial.

\section{Conclusion}

In this paper, we reviewed the various approaches to leverage monolingual data for low-resource MT. We considered methods by dividing them into Architecture Dependent and Architecture Independent categories so that it can be helpful from the practical viewpoint. We further split the methods in each of these broad categories, describing similarities and differences in their core idea, and providing available evaluation with the amount of data used. Finally, we compared both categories by improvements achieved by their methods and shared our view on what needs to be explored.

\bibliography{references}
\bibliographystyle{IEEEtranN}

\end{document}